\documentclass[11pt]{article}

\usepackage[final]{acl} 

\usepackage{times}
\usepackage{latexsym}
\usepackage[T1]{fontenc}
\usepackage[utf8]{inputenc}
\usepackage{microtype}
\usepackage{inconsolata}

\usepackage{graphicx}
\usepackage{enumitem}
\usepackage{multirow}
\usepackage{calc}
\usepackage{algorithm}
\usepackage{algorithmic} 
\usepackage{amsmath}

\title{SC-Taxo: Hierarchical Taxonomy Generation under Semantic Consistency Constraints using Large Language Models}

\author{
  Shiqiang Cai\textsuperscript{1,2}\thanks{\ \ Contributed equally to this work.} \and
  Nianhong Niu\textsuperscript{1,3}\footnotemark[1] \and
  Shizhu He\textsuperscript{1,3}\thanks{\ \ Corresponding author.} \and
  Kang Liu\textsuperscript{1,3} \and
  Jun Zhao\textsuperscript{1,3} \\
  \textsuperscript{1}Institute of Automation, Chinese Academy of Sciences, Beijing, China \\
  \textsuperscript{2}University of Science and Technology Beijing, Beijing, China \\
  \textsuperscript{3}School of Artificial Intelligence, University of Chinese Academy of Sciences, Beijing, China \\
  \texttt{shizhu.he@nlpr.ia.ac.cn}
}

\begin{document}
\maketitle

\begin{abstract}
Scientific literature is expanding at an unprecedented pace, making it increasingly challenging to efficiently organize and access domain knowledge. A high-quality scientific taxonomy offers a structured and hierarchical representation of a research field, facilitating literature exploration and topic navigation, as well as enabling downstream applications such as trend analysis, idea generation, and information retrieval. However, existing taxonomy generation approaches often suffer from structural inconsistencies and semantic misalignment across hierarchical levels. Through empirical analysis, we find that these issues largely stem from inadequate modeling of hierarchical semantic consistency. To address this limitation, we propose a semantic-consistent taxonomy generation (\textbf{SC-Taxo}) framework that leverages large language models (LLMs) with hierarchy-aware refinement stages to ensure semantic consistency. Specifically, SC-Taxo introduces a bidirectional heading generation mechanism that jointly performs bottom-up abstraction and top-down semantic constraint, while further capturing peer-level semantic dependencies to enhance horizontal consistency. Experiments on multiple benchmark datasets demonstrate consistent improvements in hierarchy alignment and heading quality, and additional evaluation on Chinese scientific literature validates its robust cross-lingual generalization.
\end{abstract}

\section{Introduction}
\begin{figure*}[!t]
\centering
\includegraphics[width=\textwidth, keepaspectratio]{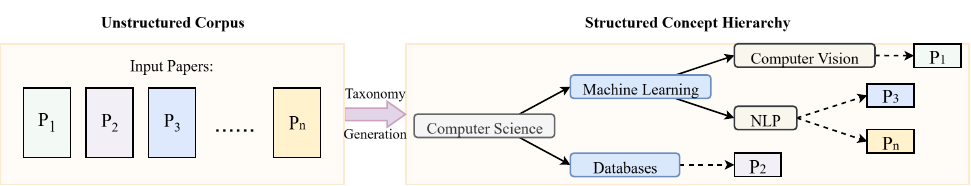}
\caption{Illustration of the scientific taxonomy generation task. The goal is to automatically organize a collection of unstructured scientific papers into a hierarchical tree of concepts, where papers are semantically mapped to the appropriate conceptual nodes at varying levels of abstraction.}
\label{figure0}
\end{figure*}

\noindent The rapid growth of scientific literature has significantly increased the demand for structured and interpretable knowledge organization systems. Hierarchical taxonomies provide a principled way to organize scientific concepts into multi-level abstractions, supporting terminology management and ontology engineering \citep{Gruber1993A,Uschold1996Ontologies}. By structuring concepts through explicit parent–child relations, taxonomies enable coherent knowledge navigation and serve as a foundational structural backbone for downstream analytical tasks such as trend analysis, idea generation, and information retrieval. As illustrated in Figure~\ref{figure0}, the core task of scientific taxonomy generation aims to automatically transform an unstructured corpus of scientific papers into such a structured concept hierarchy.

Existing taxonomy generation methods primarily fall into three categories: pattern-based extraction, unsupervised clustering, and generative approaches using large language models (LLMs). Early pattern-based methods relied on lexico-syntactic cues to extract relations \citep{Hearst1992Automatic}, while clustering algorithms group semantically related papers based on distributional similarity in embedding spaces \citep{Manning2008Introduction}. Recently, owing to their strong summarization capabilities \citep{Brown2020Language}, LLMs have been increasingly employed to directly infer conceptual hierarchies and generate expressive node headings from large corpora \citep{Hsu2024CHIME,Wan2024TnT}.

\begin{figure*}[!t]
\centering
\includegraphics[width=\textwidth, keepaspectratio]{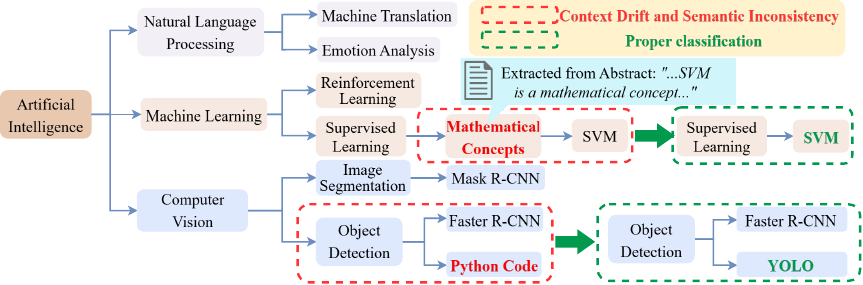} 
\caption{Contrastive analysis of semantic and structural issues in traditional taxonomy construction. Traditional methods (left) suffer from context drift—e.g., misinterpreting the local keyword ``mathematical concept'' from an abstract to generate an out-of-domain node—and granularity misalignment (e.g., placing ``Python Code'' alongside algorithms). In contrast, the proposed SC-Taxo framework (right) corrects these errors and ensures semantic coherence through explicit cross-level constraints and peer-level coordination.}
\label{figure1}
\end{figure*}

However, these approaches frequently produce hierarchies that suffer from structural distortion and semantic drift. Pattern-based methods struggle with sparse coverage and limited robustness in heterogeneous corpora. Meanwhile, clustering algorithms operate purely on proximity in embedding space rather than conceptual abstraction, and purely LLM-based methods easily suffer from semantic hallucinations without structural grounding. Consequently, the induced hierarchies often exhibit granularity misalignment and context drift. For example, as illustrated in Figure~\ref{figure1}, implementation-level items (e.g., ``Python Code'') may incorrectly appear alongside higher-level research algorithms (e.g., ``Faster R-CNN'') under a common parent node. Similarly, LLM-based generative methods can be easily hijacked by local contexts during sub-categorization. For instance, when tasked with classifying papers under the ``Supervised Learning'' node, the model processes an SVM-related paper. Upon reading the abstract phrase ``...SVM is a mathematical concept...'', the model overfits to this local keyword and loses sight of the global algorithmic context. Consequently, it erroneously generates an out-of-domain intermediate node, ``Mathematical Concepts'', directly under ``Supervised Learning'' to classify the paper, rather than correctly treating SVM as an algorithmic sub-category. This severe context drift disrupts the intended hierarchical progression.

The fundamental issue causing these structural and semantic failures is the inadequate modeling of hierarchical semantic consistency. Node relationships are often determined through shallow similarity metrics or isolated generative steps, which fail to explicitly enforce the top-down conceptual constraints and bottom-up semantic abstractions required for a coherent taxonomy. Furthermore, the lack of peer-level coordination allows redundant or conceptually overlapping categories to emerge at the same hierarchical level, severely weakening interpretability.

To address this limitation, we propose to integrate LLM-based reasoning into hierarchy-sensitive refinement stages, rather than treating language models as mere initial generators or post-hoc labeling tools. Our central premise is that semantic consistency must be explicitly constrained both vertically (across hierarchical levels) and horizontally (among sibling nodes). By decoupling the initial structural induction from deep semantic alignment, we can leverage the statistical stability of embedding-based clustering while harnessing the abstractive power of LLMs to correct semantic inconsistencies.

Based on the above motivation, we introduce \textbf{SC-Taxo (Semantic-Consistent Taxonomy Generation)}, a framework featuring a dual-path initialization and a four-round deep fusion mechanism. Specifically, SC-Taxo introduces a bidirectional heading generation mechanism that jointly performs bottom-up abstraction from paper sets and top-down semantic constraints from parent nodes. Additionally, it incorporates peer-level semantic association modeling to capture horizontal contextual relationships, thereby reducing conceptual redundancy among sibling nodes and ensuring strict granularity alignment.

Extensive experiments on the English TaxoBench benchmark and a curated Chinese scientific literature dataset demonstrate that SC-Taxo achieves consistent and significant improvements in hierarchical structural alignment and heading semantic quality. Furthermore, the framework exhibits robust cross-lingual generalization, effectively mitigating semantic collisions even in linguistically complex environments.

In summary, the main contributions of this work are three-fold:
\begin{itemize}[leftmargin=*]
    \item We propose \textbf{SC-Taxo}, a novel semantically consistent taxonomy generation framework that explicitly mitigates structural distortion and context drift by decoupling the initial topological clustering from LLM-driven semantic induction, ensuring that the generated hierarchy is grounded in both statistical structure and conceptual logic.
    \item We introduce a deep fusion mechanism that enforces holistic semantic consistency across the hierarchy; this includes bidirectional heading generation for rigorous vertical (parent-child) alignment and peer-level semantic association for horizontal (sibling) redundancy reduction and granularity balancing.
    \item We conduct comprehensive empirical evaluations on the TaxoBench benchmark and a curated Chinese scientific corpus, demonstrating that SC-Taxo significantly enhances structural isomorphism (measured by Catalogue Edit Distance Similarity, CEDS) and semantic heading quality (measured by Hierarchical Structure Retention, HSR) while exhibiting robust cross-lingual generalization.
\end{itemize}

\section{Related Work}

\subsection{Taxonomy and Ontology Learning from Text}
\noindent Constructing taxonomies from unstructured text has long relied on linguistic patterns (e.g., Hearst patterns) and graph-based modeling to extract hypernym-hyponym relations \citep{Hearst1992Automatic,Cimiano2005Learning}. Subsequent data-driven methods leveraged term co-occurrence and semantic networks to improve scalability \citep{Velardi2013OntoLearn}. To overcome the sparse coverage of rule-based systems, clustering-based techniques were adopted to organize papers into multi-level structures based on embedding similarity \citep{Manning2008Introduction}. Recent studies have further explored hybrid symbolic-statistical frameworks to balance scalability with semantic interpretability \citep{Shen2020Taxonomy,delAguilaEscobar2025Bridging}. Despite these advancements, these methods often lack explicit semantic abstraction, resulting in interpretable node descriptions that remain shallow or disconnected from the global context. Our work addresses these limitations by moving beyond pattern matching and statistical co-occurrence; SC-Taxo employs LLMs to perform deep semantic induction, ensuring that every extracted concept is grounded in both local paper content and global hierarchical constraints.

\subsection{Hierarchical Clustering for Scientific Literature}
\noindent Hierarchical clustering provides a foundational mechanism for inducing tree-structured representations from paper similarity matrices \citep{Ward1963Hierarchical}. With the emergence of neural text representations, embedding-based hierarchical clustering has become a dominant paradigm, effectively capturing coarse topical organization in large-scale scientific corpora \citep{Cohan2020SPECTER,Ayoughi2025Minimizing}. However, pure clustering-based hierarchies often fail to produce semantically meaningful internal nodes, yielding vague labels and inconsistent parent-child abstraction levels \citep{Shen2020Taxonomy}. While recent research has shifted towards semantic-guided and structure-aware clustering strategies \citep{Yang2025Semantic,Patil2025Hyperbolic}, a fundamental gap remains in maintaining structural stability while refining semantic coherence. In contrast to these approaches, SC-Taxo explicitly decouples the topological clustering path from the semantic labeling path. By treating the cluster tree as a structural scaffold, our framework allows for a four-round deep fusion process that corrects granularity misalignment and ensures consistent conceptual depth across the hierarchy.

\subsection{LLMs for Taxonomy Generation}
\noindent The rapid advancement of Large Language Models (LLMs) has enabled new paradigms for structured knowledge induction and concept labeling \citep{Zhu2023Hierarchical,Wan2024TnT}. Frameworks that incorporate LLMs into both heading generation and hierarchy refinement stages have shown promise in improving structural coherence \citep{Hsu2024CHIME,Pan2025Taxonomy,Lin2025CLIMB}. Nevertheless, treating LLMs as isolated generators or post-hoc labelers without rigid structural grounding frequently leads to ``semantic drift'' and ``hallucinated'' nodes that overfit to local abstracts. Our framework directly tackles these challenges by enforcing holistic semantic consistency through a bidirectional heading generation mechanism. Unlike previous LLM-based methods that focus on isolated generation, SC-Taxo jointly models vertical (parent-child) alignment and horizontal (sibling) peer-level associations, effectively eliminating redundancy and ensuring a compact, semantically coherent taxonomy.

\section{Materials and Methods}
\noindent The proposed SC-Taxo framework is designed to address the pervasive issues of semantic drift and structural hallucination in taxonomy generation. By decoupling structural induction from semantic generation, the framework ensures that the resulting hierarchy is grounded in both the statistical topology of the paper space and the high-level abstractive reasoning of Large Language Models.

\begin{figure*}[!t]
\centering
\includegraphics[width=\textwidth, keepaspectratio]{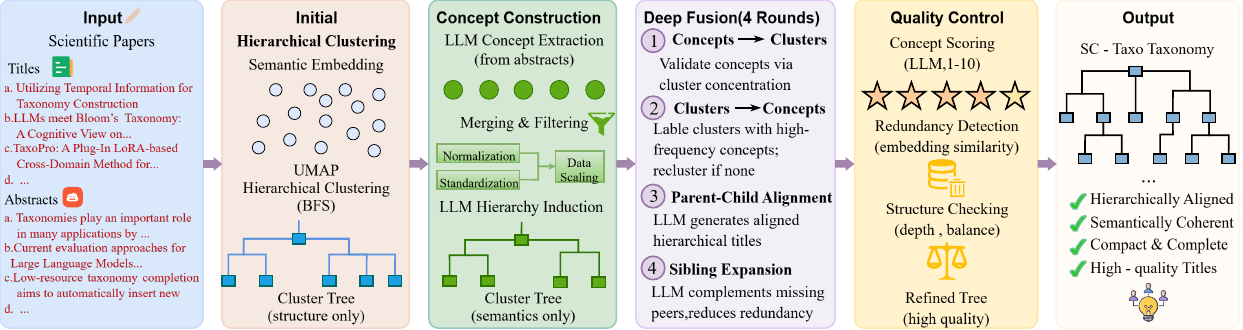} 
\caption{Overall architecture of the proposed SC-Taxo framework. The pipeline features a dual-path initialization: Initial Hierarchical Clustering constructs a structural backbone, while Concept Construction builds a semantic tree. These paths are integrated through a Deep Fusion module comprising four-round validation (Concepts $\rightarrow$ Clusters, Clusters $\rightarrow$ Concepts, Parent-Child Alignment, and Sibling Expansion). Finally, a Quality Control module applies concept scoring, redundancy detection, and structure checking to output the refined, semantically coherent taxonomy.}
\label{figure2}
\end{figure*}

As shown in Figure~\ref{figure2}, the pipeline takes a scientific corpus $D$ (comprising titles and abstracts) as input and processes it through four integrated modules: Initial Hierarchical Clustering, Concept Construction, Deep Fusion (four Rounds), and Quality Control.

\subsection{Dual-Path Initialization: Clustering and Concept Construction}
\noindent To prevent LLMs from generating ungrounded or hallucinated structures, SC-Taxo initiates the process via two parallel paths that capture complementary aspects of the domain knowledge.

\noindent \textbf{Initial Hierarchical Clustering (Structural Path):} This path constructs a reliable structural backbone for the taxonomy. Papers are first encoded into dense semantic embeddings. To preserve the non-linear local topological structure of the paper space, we apply Uniform Manifold Approximation and Projection (UMAP) for dimensionality reduction. Subsequently, a recursive Breadth-First Search (BFS) hierarchical clustering algorithm partitions the papers into a structural Cluster Tree $T_{cluster}$. While $T_{cluster}$ provides a stable scaffold, it lacks interpretable semantic labels at this stage.

\noindent \textbf{Concept Construction (Semantic Path):} In parallel, an LLM extracts key technical concepts directly from the paper abstracts. Following a phase of merging and filtering (e.g., removing low-frequency terms and normalizing synonyms), the LLM organizes these concepts into a preliminary Concept Tree $T_{concept}$. This tree is rich in semantic meaning but potentially suffers from structural instability if used as a standalone representation.

\subsection{Deep Fusion (Four-Round Validation)}
\noindent The core innovation of SC-Taxo lies in the Deep Fusion module, which aligns the structural $T_{cluster}$ with the semantic $T_{concept}$ through a rigorous four-round mutual validation process to ensure holistic hierarchical consistency.

\textbf{Round 1: Concepts $\rightarrow$ Clusters (Conceptual Grounding).} This round validates whether LLM-generated concepts are grounded in the actual paper distribution. For each concept, the framework evaluates the cluster concentration of its associated papers. Concepts whose associated papers are highly concentrated within specific clusters are retained, while overly dispersed or ``hallucinated'' concepts are filtered out.

\textbf{Round 2: Clusters $\rightarrow$ Concepts (Semantic Labeling).} Reliable semantic labels are assigned to the structural clusters by analyzing the high-frequency validated concepts within each cluster's paper subset. The most representative concept is selected as the cluster's heading. Clusters lacking clear conceptual support are flagged for recursive re-clustering to maintain structural purity.

\textbf{Round 3: Parent-Child Alignment (Vertical Consistency).} To enforce strict vertical semantic coherence, the framework refines hierarchical titles using a bidirectional mechanism. When generating a title for a child node, the LLM is conditioned on: (1) the root title (global constraint), (2) the parent node's title (local hierarchical constraint), and (3) the child node's paper subset (content grounding). This multi-level conditioning prevents the semantic drift common in unidirectional generation.

\textbf{Round 4: Sibling Expansion (Horizontal Consistency).} To enhance horizontal contextual coherence and eliminate redundancy, the framework models peer-level semantic associations. When completing a category, the LLM is provided with existing sibling topics and is explicitly prompted to identify missing technical concepts while ensuring no conceptual overlap or granularity mismatch with the established siblings.

\subsection{Quality Control and Output}
\noindent Prior to outputting the refined taxonomy $T^*$, SC-Taxo applies a rigorous Quality Control module to ensure structural compactness and semantic completeness:
\begin{itemize}[leftmargin=2em]
    \item[1.]  \textbf{Concept Scoring:} An LLM scores each concept (from 1 to 10) based on technical specificity, paper coverage, and semantic consistency. Nodes falling below a predefined quality threshold are pruned.
    \item[2.]  \textbf{Redundancy Detection:} The framework leverages embedding similarity and LLM verification to detect redundant pairs (e.g., synonyms or subsumed concepts), retaining only the most complete terminology.
    \item[3.]  \textbf{Structure Checking:} The final tree undergoes structural balancing (e.g., depth and branching factor validation) to ensure it is hierarchically aligned and semantically coherent.
\end{itemize}
The complete step-by-step workflow of the proposed SC-Taxo framework is summarized in Algorithm~\ref{alg:sc_taxo}.

\begin{algorithm}[!t]
\caption{Semantic-Consistent Taxonomy Generation (SC-Taxo)}
\label{alg:sc_taxo}
\begin{algorithmic}[1]
\REQUIRE Paper set $\mathcal{D}$, Pretrained Encoder $Enc$, LLM generative functions
\ENSURE Refined Taxonomy Tree $T^*$

\STATE \textbf{Phase 1: Dual-Path Initialization}
\STATE $E \leftarrow Enc(\mathcal{D})$
\STATE $T_{cluster} \leftarrow \text{UMAP\_BFS\_Clustering}(E)$ \COMMENT{Structural Path}
\STATE $C \leftarrow \text{LLM\_Extract\_Concepts}(\mathcal{D})$
\STATE $T_{concept} \leftarrow \text{LLM\_Induce\_Hierarchy}(C)$ \COMMENT{Semantic Path}

\STATE \textbf{Phase 2: Deep Fusion (Four-Round Validation)}
\STATE $C_{valid} \leftarrow \text{Validate\_Concepts\_via\_Clusters}(C, T_{cluster})$ \COMMENT{Round 1}
\STATE $T_{labeled} \leftarrow \text{Label\_Clusters\_with\_Concepts}(T_{cluster}, C_{valid})$ \COMMENT{Round 2}
\STATE $T_{aligned} \leftarrow \text{Bidirectional\_Alignment}(T_{labeled})$ \COMMENT{Round 3: Parent-Child}
\STATE $T_{expanded} \leftarrow \text{Expand\_Siblings}(T_{aligned})$ \COMMENT{Round 4: Peer-level}

\STATE \textbf{Phase 3: Quality Control}
\STATE $T_{scored} \leftarrow \text{Score\_Concepts}(T_{expanded})$
\STATE $T^* \leftarrow \text{Remove\_Redundancy}(T_{scored})$
\RETURN $T^*$
\end{algorithmic}
\end{algorithm}

\subsection{Dataset and Benchmark}
\noindent We evaluate the proposed framework on the TaxoBench benchmark, which provides expert-annotated hierarchical taxonomies for scientific literature collections \citep{Zhu2023Hierarchical}. Each benchmark instance consists of a survey paper, its referenced papers, and a manually curated taxonomy structure. Following standard practice, papers are represented using their titles and abstracts.

To assess cross-lingual generalization, we additionally curated a Chinese scientific literature dataset derived from a comprehensive domain survey focusing on the intersection of Large Language Models (LLMs) and Reinforcement Learning (RL) agents. This dataset comprises 59 research papers, with the ground-truth hierarchical taxonomy expertly annotated based on the original survey's structure. Compared to English corpora, Chinese scientific literature often exhibits implicit topic boundaries, higher semantic ambiguity, and less standardized terminology, making taxonomy generation more challenging \citep{Ruder2019A}. The annotation schema follows the principles of TaxoBench to ensure comparability, and identical evaluation metrics are applied across languages.

\subsection{Implementation Details and Reproducibility}
\noindent All large language model interactions in our framework are implemented through prompt-based generation and evaluation. To avoid introducing additional variability, the same base LLM is used consistently across heading generation, semantic alignment, and refinement stages.

For paper representation, paper titles and abstracts are concatenated and encoded using a fixed pre-trained embedding model. The resulting dense embeddings are used exclusively for the initial hierarchical clustering stage and are not updated during subsequent processing.

During bidirectional heading generation, bottom-up abstraction and top-down refinement are each performed once per node, following a strictly single bottom-up pass and a single top-down pass through the taxonomy. Parent–child semantic consistency is evaluated using soft alignment criteria, implemented via LLM-based textual scoring or embedding-based similarity, depending on availability.

Peer-level semantic associations are computed only among sibling nodes at the same hierarchy level and are used as auxiliary semantic signals during refinement. Crucially, they do not alter the foundational hierarchical structure induced by the clustering path.

All hyperparameters, including the UMAP dimensionality, BFS clustering depth, and LLM validation thresholds, are kept strictly fixed across all experiments. Furthermore, the identical prompting templates and evaluation criteria are applied to both the English and Chinese datasets to ensure fair cross-lingual comparison and full reproducibility.

\section{Results}
\subsection{Evaluation Metrics}
\noindent We evaluate taxonomy quality using a comprehensive set of both flat clustering and hierarchy-aware metrics. To assess paper-level clustering consistency, we employ Normalized Mutual Information (NMI) and Purity.

To rigorously evaluate the hierarchical structural alignment and semantic quality, we utilize two specialized metrics introduced in the TaxoBench framework \citep{Zhu2023Hierarchical}:

\textbf{Catalogue Edit Distance Similarity (CEDS):} This metric measures the structural and semantic isomorphism between the predicted taxonomy $T_{pred}$ and the ground truth $T_{gt}$. It computes the minimum tree edit distance—allowing node insertion, deletion, and semantic-aware substitution—strictly normalized by the maximum tree size.

\textbf{Hierarchical Structure Retention (HSR):} HSR evaluates how well the ancestral relationships among papers are preserved. It computes the symmetric bounded ratio of the Lowest Common Ancestor (LCA) depths for paper pairs in $T_{pred}$ compared to their corresponding depths in $T_{gt}$.

Together, CEDS and HSR explicitly quantify the improvements in hierarchical coherence and abstraction depth achieved by our proposed framework.

\subsection{Compared Baselines}
\noindent To demonstrate the superiority of SC-Taxo, we compare it against a diverse set of state-of-the-art taxonomy generation and clustering methods, categorized into three groups:

\begin{itemize}
    \item \textbf{Pure LLM-based Methods}: Methods that directly prompt LLMs to generate taxonomies or clusters, including CHIME \citep{Hsu2024CHIME} (which iteratively abstracts concepts) and TnT-LLM \citep{Wan2024TnT} / GoalEx \citep{GoalEx_Citation} (which generate goal-driven hierarchical clusters based on textual representations).
    \item \textbf{Clustering-incorporated Methods}: Approaches that use traditional clustering as a structural backbone. We compare against Knowledge Navigator (KN) \citep{Katz2024Knowledge} and SCYCHIC \citep{SCYCHIC_Citation}, which leverage paper embeddings for structural induction prior to label assignment.
    \item \textbf{LLM-Guided Multi-Aspect Clustering}: The most competitive recent baselines, such as CAM-Taxo \citep{Zhu2025Context}, which integrate LLM reasoning to guide the clustering process along specific semantic dimensions.
\end{itemize}

\subsection{Main Results on English TaxoBench Benchmark}

\begin{table}[!t]
  \centering
  \caption{Performance comparison of different taxonomy generation methods on the TaxoBench benchmark.}
  \label{tab:english_results}
  \resizebox{\linewidth}{!}{
  \begin{tabular}{lcccc} 
    \hline
    \multirow{2}{*}{\textbf{Method}} & \multicolumn{2}{c}{\textbf{Categorization}} & \multicolumn{2}{c}{\textbf{Structure}} \\
    \cline{2-5}
     & NMI & Purity & CEDS & HSR \\  
    \hline
    \textbf{Pure LLM-based} & & & & \\
    CHIME & 35.4 & 41.8 & 23.3 & 74.7 \\
    TnT-LLM & 51.6 & 57.6 & 19.1 & 69.9 \\
    GoalEx & 46.7 & 47.6 & 23.2 & 70.5 \\
    \hline
    \textbf{Clustering-incorporated} & & & & \\
    KN & 44.7 & 42.4 & 18.8 & 49.5 \\
    SCYCHIC & 49.8 & 50.6 & 23.0 & 66.4 \\
    \hline
    \multicolumn{5}{l}{\textbf{LLM-Guided Multi-Aspect Clustering}} \\
    CAM-Taxo & \textbf{60.1} & 62.2 & 23.8 & 74.5 \\
    \hline
    SC-Taxo (Ours) & 50.2 & \textbf{64.8} & \textbf{28.7} & \textbf{75.3} \\ 
    \hline
  \end{tabular}
  }
\end{table}

\noindent As shown in Table~\ref{tab:english_results}, the results on the TaxoBench benchmark demonstrate that the proposed SC-Taxo framework significantly outperforms existing baselines, particularly in structural alignment (CEDS) and paper categorization quality (Purity). 

When analyzing the categorization metrics, SC-Taxo achieves the highest Purity score (64.8), surpassing even the highly competitive CAM-Taxo (62.2). This substantial gain indicates that our Deep Fusion module—specifically the mechanism that validates LLM-extracted concepts against actual cluster concentrations—effectively filters out hallucinated or overly broad topics, ensuring that papers assigned to a specific node share a highly cohesive semantic focus. Interestingly, while our NMI (50.2) is slightly lower than that of CAM-Taxo (60.1), this is a deliberate trade-off. NMI strictly evaluates flat paper partitioning, whereas SC-Taxo is optimized for hierarchical abstraction. By sacrificing a minor degree of flat clustering agreement, our framework avoids over-fragmenting papers into granular, semantically meaningless sub-clusters, which ultimately benefits the global taxonomy structure.

The most striking improvements are observed in the hierarchy-aware metrics. SC-Taxo establishes a new state-of-the-art CEDS score of 28.7, outperforming the best baseline by nearly 5 absolute percentage points. Pure LLM-based methods like CHIME and TnT-LLM struggle to construct stable global trees (CEDS around 19-23) because they lack topological grounding. Conversely, clustering-incorporated methods like KN suffer from weak label semantics (HSR: 49.5). Our dual-path architecture bridges this gap: the bidirectional heading generation acts as a strict semantic regularizer, ensuring that child nodes strictly inherit constraints from parent nodes while accurately summarizing their own papers. Consequently, SC-Taxo not only generates more semantically reliable abstractions (HSR: 75.3) but also ensures that the overall tree isomorphism is highly consistent with human-curated expert taxonomies.

\subsection{Cross-lingual Evaluation on Chinese Scientific Literature}
\noindent Evaluating taxonomy generation across languages is critical for assessing whether a framework models deep semantic structure rather than relying on language-specific surface lexical patterns. Scientific literature in Chinese poses unique and severe challenges, including implicit topic boundaries, highly compound terminological variability, and weaker standardization compared to English corpora. 

\begin{table}[!t] 
  \centering
  \caption{Performance comparison on the Chinese TaxoBench benchmark.}
  \label{tab:chinese_results}
  \resizebox{\linewidth}{!}{
  \begin{tabular}{lcccc} 
    \hline
    \multirow{2}{*}{\textbf{Method}} & \multicolumn{2}{c}{\textbf{Categorization}} & \multicolumn{2}{c}{\textbf{Structure}} \\
    \cline{2-5}
     & NMI & Purity & CEDS & HSR \\  
    \hline
    \textbf{Pure LLM-based} & & & & \\
    GoalEx & 30.3 & 38.1 & 14.8 & 23.5 \\
    \hline
    \textbf{Clustering-incorporated} & & & & \\
    KN & 25.9 & 32.2 & 21.5 & 46.2 \\
    \hline
    \multicolumn{5}{l}{\textbf{LLM-Guided Multi-Aspect Clustering}} \\
    CAM-Taxo & \textbf{52.2} & 55.9 & 20.6 & 50.2 \\
    \hline
    SC-Taxo (Ours) & 30.4 & \textbf{66.7} & \textbf{29.8} & \textbf{51.5} \\ 
    \hline
  \end{tabular}
  }
\end{table}

To rigorously evaluate performance under these challenging zero-shot conditions, we conducted experiments on a curated Chinese dataset focused on the intersection of Large Language Models (LLMs) and Reinforcement Learning (RL) agents. The ground-truth taxonomy for this dataset was expertly annotated based on a comprehensive domain survey, encompassing a complex hierarchy of tasks such as reward shaping, state representation, and policy optimization. 

Table~\ref{tab:chinese_results} presents the quantitative comparison between our SC-Taxo framework and competitive baselines on this Chinese corpus. The results reveal a catastrophic degradation in traditional generative methods. For instance, GoalEx struggles immensely with the terminological ambiguity of Chinese text, yielding a heavily fragmented tree with a remarkably low CEDS score of 14.8 and an HSR of 23.5. This highlights the vulnerability of pure generative pipelines when applied to non-English scientific domains where standard prompt formulations lose their structural efficacy.

In stark contrast, the proposed SC-Taxo framework maintains highly robust performance across the language barrier, achieving state-of-the-art structural alignment (CEDS: 29.8) and preserving high semantic heading accuracy (HSR: 51.5). The reason for this robust cross-lingual generalization lies in our decoupled architecture. The UMAP+BFS structural path provides a language-agnostic topological anchor based purely on dense embeddings, preventing the hierarchy from collapsing. Meanwhile, the peer-level semantic association module explicitly checks for conceptual redundancy, effectively resolving the synonym collisions that frequently occur in Chinese terminology. Furthermore, achieving the highest Purity (66.7) among all baselines confirms that our multi-round deep fusion mechanism successfully mitigates context drift, proving that SC-Taxo captures universal conceptual regularities rather than overfitting to English-centric syntax.

\subsection{Ablation Study on Framework Components}
\noindent To rigorously evaluate the contribution of each core component in the SC-Taxo framework, we conduct a component-wise ablation study on the TaxoBench benchmark. Rather than treating the framework as a monolithic black box, this ablation analysis systematically dissects the synergistic effects between the structural backbone and the LLM-driven semantic refinement modules. The evaluation focuses on three key metrics:
\begin{itemize}[noitemsep]
    \item \textbf{CEDS}, measuring hierarchical structural alignment;
    \item \textbf{HSR}, reflecting heading semantic accuracy;
    \item \textbf{Purity}, indicating paper-level clustering consistency.
\end{itemize}

We compare the full SC-Taxo model against four distinct variants to isolate the impact of each specific design choice:
\begin{enumerate}[noitemsep]
    \item \textbf{BU-only (Bottom-Up only):} Retains only the structural clustering path, completely disabling the top-down concept extraction and all subsequent fusion steps;
    \item \textbf{w/o Bi (w/o Bidirectional Validation):} Retains both the clustering and concept paths but removes the bidirectional validation mechanism;
    \item \textbf{w/o Peer (w/o Sibling Expansion):} Disables the horizontal semantic expansion among peer nodes;
    \item \textbf{w/o Refine (w/o Quality Filtering):} Removes the final LLM-driven quality scoring and redundancy elimination module.
\end{enumerate}

\begin{figure}[!t]
\centering
\includegraphics[width=\linewidth, keepaspectratio, clip, trim=0 10 0 10]{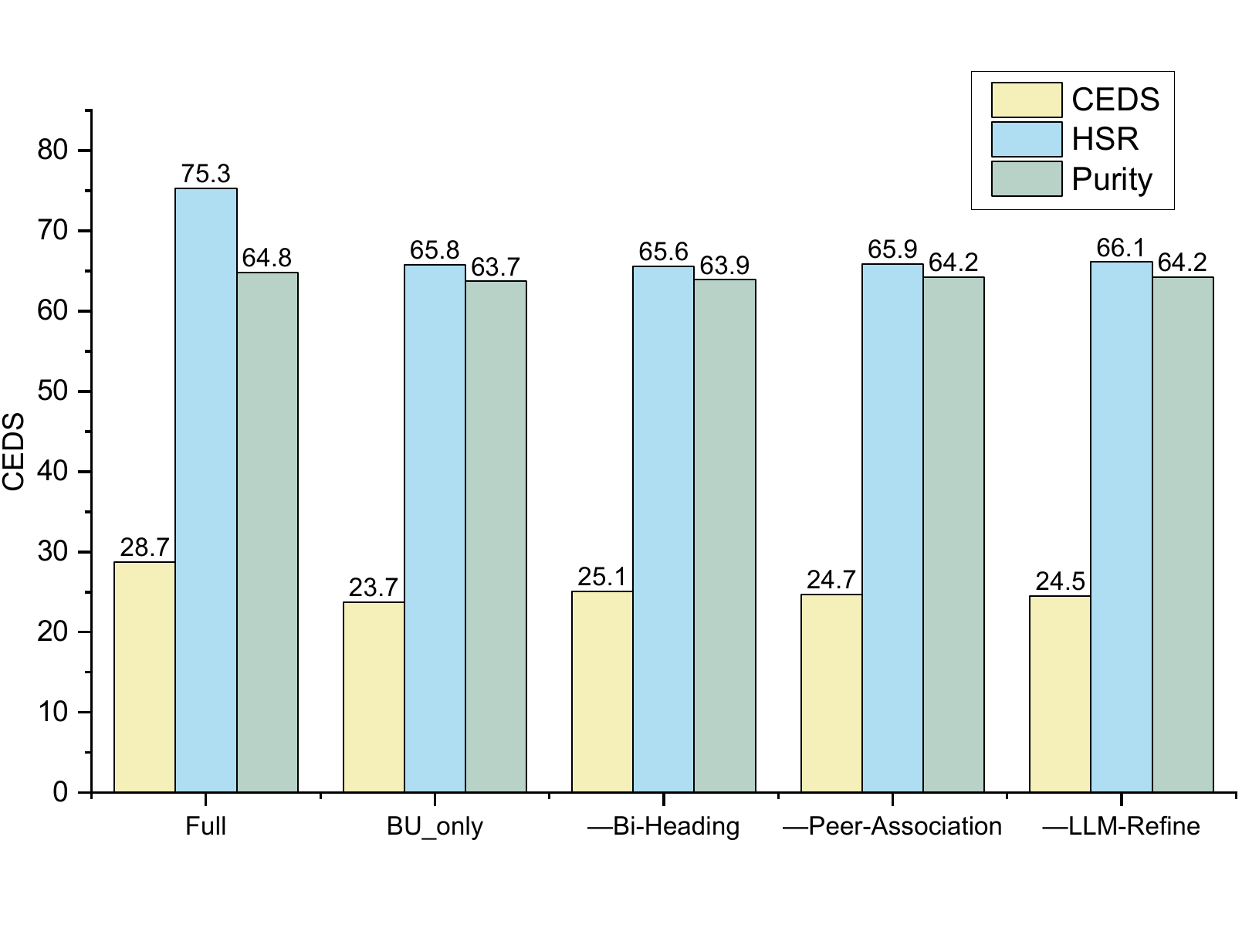}
\caption{Ablation results on the TaxoBench benchmark. The substantial performance gap between BU-only, w/o Bi, and the Full model highlights the critical necessity of both the dual-path architecture and the bidirectional validation mechanism.}
\label{figure3}
\end{figure}

The evaluation results across structural and semantic metrics are presented in Figure~\ref{figure3}. Overall, the full SC-Taxo framework achieves the best performance across all metrics (e.g., CEDS: 28.7, HSR: 75.3, Purity: 64.8), validating the effectiveness of our holistic dual-path design.

\subsubsection{The Necessity of Dual-Path and Bidirectional Validation}
\noindent The \textbf{BU-only} variant, which completely removes the top-down LLM concept path and relies solely on structural paper features, yields the lowest overall performance (CEDS drops to 23.7, Purity to 63.7). This confirms that traditional bottom-up clustering struggles to capture high-level semantic abstractions. When the concept path is introduced but without bidirectional validation (\textbf{w/o Bi}), the semantic alignment (CEDS) slightly improves to 25.1. However, it still falls significantly short of the full model (28.7). This substantial performance gap (28.7 vs. 25.1) is crucial: it demonstrates that merely injecting LLM concepts is insufficient. Without bidirectional validation acting as a bridge to resolve the heterogeneity between structural clusters and semantic concepts, the taxonomy suffers from semantic collisions. The deep fusion mechanism is therefore essential for ensuring hierarchical consistency.

\subsubsection{Impact of Quality Refinement} 
\noindent Removing the LLM-driven refinement module (\textbf{w/o Refine}) causes a sharp decline in CEDS (from 28.7 to 24.5) and Purity (from 64.8 to 64.2). This empirical evidence confirms that our post-processing step effectively filters out noisy labels and reduces conceptual redundancy, ensuring strict semantic consistency between the generated headings and their constituent papers.

\subsubsection{Impact of Peer-level Expansion} 
\noindent Finally, disabling the horizontal sibling expansion (\textbf{w/o Peer}) leads to a notable decrease in CEDS (to 24.7) and HSR (from 75.3 to 65.9). This highlights that explicitly modeling peer-level associations is critical for discovering latent relationships among sibling nodes, thereby maximizing the semantic coverage and structural completeness of the taxonomy.

\subsection{Case Study: Qualitative Analysis}

\begin{figure*}[!t]
    \centering
    \includegraphics[width=\textwidth, keepaspectratio]{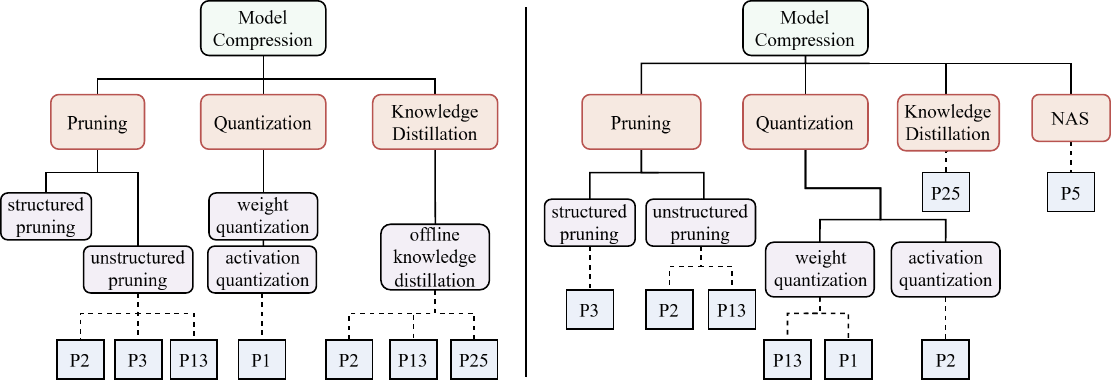} 
    \vspace{2mm} 
    \begin{minipage}[t]{0.48\textwidth}
        \centering
        (a) Generated by \citep{Zhu2025Context}
    \end{minipage}
    \hfill 
    \begin{minipage}[t]{0.48\textwidth}
        \centering
        (b) Generated by SC-Taxo (Ours)
    \end{minipage}   
    \vspace{4mm} 
    \caption{Qualitative comparison of taxonomies generated for the topic ``Model Compression''. (a) The baseline method suffers from unbalanced granularity. (b) Our proposed SC-Taxo ensures structural compactness by collapsing redundant levels and corrects paper assignments.}
    \label{figure4}
\end{figure*}

\noindent To intuitively demonstrate the superiority of our proposed framework over existing LLM-based taxonomy generation methods, we conduct a qualitative case study on a subset of papers related to ``Model Compression''. Figure~\ref{figure4} visualizes the taxonomy sub-trees generated by the baseline method and our SC-Taxo.

As shown in Figure~\ref{figure4}(a), the baseline approach exhibits granularity mismatch and structural imbalance. On one hand, it generates empty categories without paper support (e.g., ``structured pruning'' and ``weight quantization'') while improperly lumping distinct papers together (e.g., misclassifying P3 alongside P2 and P13 under ``unstructured pruning''). On the other hand, it suffers from over-fragmentation by creating a redundant single-child node (``offline knowledge distillation''), making the taxonomy unnecessarily deep and complex.

In contrast, Figure~\ref{figure4}(b) illustrates the refined taxonomy generated by SC-Taxo. Driven by the four-round Deep Fusion and Quality Control modules, our framework successfully resolves these issues. First, it ensures strict semantic alignment by correctly differentiating paper contexts, moving P3 into ``structured pruning'' and accurately populating ``weight quantization'' with relevant papers (P1 and P13). Second, it mitigates over-fragmentation by collapsing the redundant ``offline knowledge distillation'' node, assigning P25 directly to ``Knowledge Distillation''. Finally, through peer-level semantic association, SC-Taxo successfully identifies and expands a critical missing sibling category, ``NAS'' (Neural Architecture Search), resulting in a more comprehensive, compact, and structurally balanced taxonomy.

\subsection{Analysis of Key Metrics}
\noindent As shown in the experimental results, the proposed framework achieves higher scores on Purity and CEDS compared to baseline methods. In particular, the improvement in CEDS indicates that the generated taxonomy exhibits better global structural alignment with the reference hierarchy. This suggests that explicitly modeling parent–child semantic consistency contributes to more coherent hierarchical organization, even when the underlying paper clustering is not further optimized.

The framework also shows a modest but consistent improvement in HSR. Although the absolute gain is limited, the increase indicates that the generated headings are slightly closer to the ground-truth concepts. This suggests that the refinement and filtering stage improves semantic precision without introducing aggressive abstraction, thereby favoring stable and reliable heading generation over marginal lexical optimization.

By contrast, NMI scores are lower than those of some baseline methods. Since this metric primarily reflects paper-level clustering agreement, the decrease indicates that the proposed framework does not prioritize optimizing flat clustering consistency. Instead, the method deliberately trades off paper-level clustering performance for improved hierarchical coherence and semantic interpretability, which are more critical for conceptual taxonomy construction.

Overall, the results demonstrate that improvements in hierarchical alignment and semantic consistency can be achieved without optimizing traditional clustering metrics, highlighting the practical effectiveness of the proposed framework for constructing interpretable conceptual hierarchies.

\section{Discussion}
\noindent The experimental results indicate that semantic consistency, rather than clustering optimization alone, plays a decisive role in constructing interpretable conceptual hierarchies. By integrating bidirectional heading alignment and peer-level semantic modeling, the proposed framework reduces hierarchical semantic drift while maintaining structural stability \citep{Mishra2025QuanTaxo,Patil2025Hyperbolic}. The ablation analysis further confirms that different modules contribute complementary improvements to alignment and semantic precision.

A remaining limitation lies in the computational overhead introduced by multi-stage LLM invocation. Future work may explore lightweight consistency estimation strategies or adaptive refinement mechanisms to balance efficiency and semantic robustness.

\section{Conclusion}
\noindent This study presents SC-Taxo, a semantic-consistent framework for hierarchy taxonomy generation in scientific literature. By decoupling structural induction from semantic refinement, the framework combines embedding-based hierarchical clustering with bidirectional heading alignment, peer-level association modeling, and LLM-driven validation. Experimental results demonstrate improved hierarchical alignment and heading semantic quality, particularly in CEDS and HSR, while maintaining stable clustering behavior. Cross-lingual evaluation further shows that the framework generalizes across languages, indicating that it captures language-agnostic conceptual regularities. The findings highlight the importance of explicitly modeling hierarchical semantics in taxonomy construction. Future work may extend the framework to more flexible graph structures and explore efficiency-aware semantic refinement strategies.

\section*{Conflicts of Interest}
\noindent The authors declare that they have no conflict of interest.



\begin{thebibliography}{99}

\bibitem[Artetxe \& Schwenk, 2019]{Artetxe2019Massively}
Artetxe M, Schwenk H.
Massively multilingual sentence embeddings for zero-shot cross-lingual transfer and beyond.
{\it Transactions of the Association for Computational Linguistics}, 2019; {\bf 7}: 597–610.

\bibitem[Ayoughi et al., 2025]{Ayoughi2025Minimizing}
Ayoughi M, Mettes P, Groth P. 
Minimizing Hyperbolic Embedding Distortion with LLM-Guided Hierarchy Restructuring.
In: {\it Proceedings of the 13th Knowledge Capture Conference 2025}, 2025: 123–130.

\bibitem[Buitelaar et al., 2005]{Buitelaar2005Ontology}
Buitelaar P, Cimiano P, Magnini B. 
Ontology learning from text: Methods, evaluation and applications. 
{\it IOS Press}, 2005.

\bibitem[Brown et al., 2020]{Brown2020Language}
Brown T B, Mann B, Ryder N, et al.
Language models are few-shot learners.
{\it Advances in Neural Information Processing Systems (NeurIPS)}, 2020; {\bf 33}: 1877–1901.

\bibitem[Chen et al., 2024]{Chen2024SAC}
Chen H, Shen X, Lv Q, Wang J, Ni X, Ye J.
SAC-KG: Exploiting Large Language Models as Skilled Automatic Constructors for Domain Knowledge Graph.
In: {\it Proceedings of the 62nd Annual Meeting of the Association for Computational Linguistics (ACL 2024)}, 2024: 4345–4360.

\bibitem[Cimiano et al., 2005]{Cimiano2005Learning}
Cimiano P, Hotho A, Staab S. 
Learning concept hierarchies from text corpora using formal concept analysis. 
{\it Journal of Artificial Intelligence Research}, 2005; {\bf 24}: 305–339.

\bibitem[Cimiano, 2006]{Cimiano2006Ontology}
Cimiano P. 
Ontology Learning from Text. 
{\it Springer}, 2006.

\bibitem[Cohan et al., 2020]{Cohan2020SPECTER}
Cohan A, Feldman S, Beltagy I, Downey D, Weld D S.
SPECTER: Document-level representation learning using citation-informed transformers.
In: {\it Proceedings of the 58th Annual Meeting of the Association for Computational Linguistics}, 2020: 2270–2282.

\bibitem[del Águila Escobar et al., 2025]{delAguilaEscobar2025Bridging}
del Águila Escobar R A, del Carmen Suárez-Figueroa M, Fernández López M, Villazón Terrazas B.
Bridging Text and Knowledge: Explainable AI for Knowledge Graph Classification and Concept Map-Based Semantic Domain Discovery with OBOE Framework.
{\it Appl. Sci.}, 2025; {\bf 15}(22): 12231.

\bibitem[Fortuna \& Lavrač, 2009]{Fortuna2009Ontology}
Fortuna B, Lavrač N.
Ontology learning from text: A survey.
{\it Knowledge and Information Systems}, 2009; {\bf 19}(1): 1–35.

\bibitem[Gao et al., 2025]{SCYCHIC_Citation}
Gao M, Shah J, Wang W, Khashabi D.
Science Hierarchography: Hierarchical Organization of Science Literature.
{\it arXiv preprint arXiv:2504.13834}, 2025.

\bibitem[Gruber, 1993]{Gruber1993A}
Gruber TR. 
A translation approach to portable ontology specifications. 
{\it Knowledge Acquisition}, 1993; {\bf 5}(2): 199–220.

\bibitem[Hearst, 1992]{Hearst1992Automatic}
Hearst M A.
Automatic acquisition of hyponyms from large text corpora.
In: {\it Proceedings of the 14th International Conference on Computational Linguistics (COLING)}, 1992: 539–545.

\bibitem[Hsu et al., 2024]{Hsu2024CHIME}
Hsu C-C, Bransom E, Sparks J, et al. 
CHIME: LLM-assisted hierarchical organization of scientific studies for literature review support. 
In: {\it Findings of the Association for Computational Linguistics (ACL)}, 2024: 118–132.

\bibitem[Ilman et al., 2025]{Ilman2025CUET}
Ilman R, Rahman M, Rahman S. 
CUET\_Zenith at LLMs4OL 2025 Task C: Hybrid Embedding-LLM Architectures for Taxonomy Discovery.
{\it Open Conference Proceedings}, 2025; {\bf 6}: 2896. 

\bibitem[Jamatia et al., 2025]{Jamatia2025TaxoAdapt}
Jamatia A, Mitra P, Hovy E H.
TaxoAdapt: Aligning LLM-Based Multidimensional Taxonomy Construction to Evolving Research Corpora.
In: {\it Proceedings of the 63rd Annual Meeting of the Association for Computational Linguistics (Volume 1: Long Papers)}, 2025: 25911–25928.

\bibitem[Ji et al., 2022]{Ji2022A}
Ji S, Pan S, Cambria E, et al. 
A survey on knowledge graphs: Representation, acquisition, and applications. 
{\it IEEE Transactions on Neural Networks and Learning Systems}, 2022; {\bf 33}(2): 494–514.

\bibitem[Katz et al., 2024]{Katz2024Knowledge}
Katz U, Levy M, Goldberg Y. 
Knowledge Navigator: LLM-guided browsing framework for exploratory search in scientific literature. 
In: {\it Findings of the Conference on Empirical Methods in Natural Language Processing (EMNLP)}, 2024: 8838–8855.

\bibitem[Lin et al., 2025]{Lin2025CLIMB}
Lin N, et al.
Building Data-Driven Occupation Taxonomies: A Bottom-Up Multi-Stage Approach via Semantic Clustering and Multi-Agent Collaboration.
In: {\it Proceedings of the 2025 Conference on Empirical Methods in Natural Language Processing: Industry Track (EMNLP 2025)}, 2025.

\bibitem[Maedche \& Staab, 2001]{Maedche2001Ontology}
Maedche A, Staab S. 
Ontology learning for the semantic web. 
{\it IEEE Intelligent Systems}, 2001; {\bf 16}(2): 72–79.

\bibitem[Manning et al., 2008]{Manning2008Introduction}
Manning C D, Raghavan P, Schütze H.
Introduction to Information Retrieval.
{\it Cambridge University Press}, 2008.

\bibitem[Mishra et al., 2025]{Mishra2025QuanTaxo}
Mishra S, Patni A, Chatterjee N, Chakraborty T.
QuanTaxo: A quantum approach to self-supervised taxonomy expansion.
{\it arXiv preprint arXiv:2501.14011}, 2025.

\bibitem[Navigli et al., 2011]{Navigli2011A}
Navigli R, Velardi P, Faralli S. 
A graph-based algorithm for inducing lexical taxonomies from scratch. 
In: {\it Proceedings of the International Joint Conference on Artificial Intelligence (IJCAI)}, 2011: 1872–1877.

\bibitem[Patil et al., 2025]{Patil2025Hyperbolic}
Patil S, Zhang Z, Huang Y, Ma T, Xu M.
Hyperbolic large language models.
{\it arXiv preprint arXiv:2509.05757}, 2025.

\bibitem[Pan et al., 2025]{Pan2025Taxonomy}
Pan H, Zhang Q, Adamu M, Dragut E, Latecki L J.
Taxonomy-Driven Knowledge Graph Construction for Domain-Specific Scientific Applications.
In: {\it Findings of the Association for Computational Linguistics: ACL 2025}, 2025: 4295–4320.

\bibitem[Reimers \& Gurevych, 2019]{Reimers2019Sentence}
Reimers N, Gurevych I.
Sentence-BERT: Sentence embeddings using Siamese BERT-networks.
In: {\it Proceedings of the 2019 Conference on Empirical Methods in Natural Language Processing and the 9th International Joint Conference on Natural Language Processing (EMNLP-IJCNLP)}, 2019: 3982–3992.

\bibitem[Ruder et al., 2019]{Ruder2019A}
Ruder S, Vulić I, Søgaard A. 
A survey of cross-lingual word embedding models. 
{\it Journal of Artificial Intelligence Research}, 2019; {\bf 65}: 569–631.

\bibitem[Shen et al., 2020]{Shen2020Taxonomy}
Shen J, Ma X, Zhang H, Li J.
Taxonomy construction for scientific literature.
{\it Knowledge-Based Systems}, 2020; {\bf 191}: 105240.

\bibitem[Tabatabaei et al., 2025]{Tabatabaei2025LLMClass}
Tabatabaei S A, Fancher S, Parsons M, Askari A.
Can Large Language Models Serve as Effective Classifiers for Hierarchical Multi-Label Classification of Scientific Papers at Industrial Scale?
In: {\it Proceedings of the 2025 International Conference on Computational Linguistics (COLING 2025)}, 2025: 167–179.

\bibitem[Uschold \& Grüninger, 1996]{Uschold1996Ontologies}
Uschold M, Grüninger M. 
Ontologies: Principles, methods and applications. 
{\it Knowledge Engineering Review}, 1996; {\bf 11}(2): 93–136.

\bibitem[Velardi et al., 2013]{Velardi2013OntoLearn}
Velardi P, Faralli S, Navigli R. 
OntoLearn Reloaded: A graph-based algorithm for taxonomy induction. 
{\it Computational Linguistics}, 2013; {\bf 39}(3): 665–707.

\bibitem[Vu et al., 2025]{Vu2025Automated}
Vu B, Naik R G, Nguyen B K, Mehraeen S, Hemmje M.
Automated taxonomy construction using large language models: A comparative study of fine-tuning and prompt engineering.
{\it Eng}, 2025; {\bf 6}(11): 283.

\bibitem[Wan et al., 2024]{Wan2024TnT}
Wan M, Safavi T, Jauhar SK, et al. 
TnT-LLM: Text mining at scale with large language models. 
In: {\it Proceedings of the ACM SIGKDD Conference on Knowledge Discovery and Data Mining (KDD)}, 2024: 5836–5847.

\bibitem[Wang et al., 2023]{GoalEx_Citation}
Wang Z, Shang J, Zhong R.
Goal-driven explainable clustering via language descriptions.
In: {\it Proceedings of the 2023 Conference on Empirical Methods in Natural Language Processing (EMNLP)}, 2023: 10626–10649.

\bibitem[Ward, 1963]{Ward1963Hierarchical}
Ward J H.
Hierarchical Grouping to Optimize an Objective Function.
{\it Journal of the American Statistical Association}, 1963; {\bf 58}(301): 236–244.

\bibitem[Yang et al., 2025]{Yang2025Semantic}
Yang D, Zhang X, Han T, Liu Y.
Semantic enrichment of neural word embeddings: Leveraging taxonomic similarity for enhanced distributional semantics.
{\it Natural Language Processing}, 2025; {\bf 31}(6): 1423–1449.

\bibitem[Zhang et al., 2025a]{Zhang2025LLMTaxo}
Zhang H, Zhu Z, Zhang Z, Li C.
LLMTaxo: Leveraging Large Language Models for Constructing Taxonomy of Factual Claims from Social Media.
In: {\it Findings of the Association for Computational Linguistics: ACL 2025}, 2025: 19627–19641.

\bibitem[Zang et al., 2025]{Zang2025KG}
Zang Q, Zgrzendek C, Tchappi I, Khadangi A, Sedlmeir J.
KG-HTC: Integrating knowledge graphs into LLMs for effective zero-shot hierarchical text classification.
{\it arXiv preprint arXiv:2505.05583}, 2025.

\bibitem[Zhang et al., 2021]{Zhang2021Hierarchical}
Zhang Y, Chen X, Meng Y, et al. 
Hierarchical metadata-aware document categorization under weak supervision. 
In: {\it Proceedings of the ACM International Conference on Web Search and Data Mining (WSDM)}, 2021: 770–778.

\bibitem[Zhang et al., 2025b]{Zhang2025Construction}
Zhang Y, Xu W, Yu Z, Reformat M Z.
Construction of topic hierarchy with subtree representation for knowledge graphs.
{\it Axioms}, 2025; {\bf 14}(4): 300.

\bibitem[Zhu et al., 2023]{Zhu2023Hierarchical}
Zhu K, Feng X, Feng X, et al. 
Hierarchical catalogue generation for literature review: A benchmark. 
In: {\it Findings of the Conference on Empirical Methods in Natural Language Processing (EMNLP)}, 2023: 6790–6804.

\bibitem[Zhu et al., 2025]{Zhu2025Context}
Zhu K, Liao L, Gu Y, et al.
Context-aware hierarchical taxonomy generation for scientific papers via LLM-guided multi-aspect clustering.
In: {\it Proceedings of the 2025 Conference on Empirical Methods in Natural Language Processing (EMNLP)}, 2025: 15627–15645.

\end{thebibliography}
\end{document}